\documentclass{article}

\usepackage[preprint]{neurips_2021}

\usepackage[utf8]{inputenc} 
\usepackage[T1]{fontenc}    
\usepackage{hyperref}       
\usepackage{url}            
\usepackage{booktabs}       
\usepackage{amsfonts,amssymb,amsmath}       
\usepackage{nicefrac}       
\usepackage{microtype}      
\usepackage{xcolor}         
\usepackage{multicol}
\usepackage{multirow}
\usepackage{graphicx}

\newcommand{\bcket}[3]{\left#1 #3 \right#2}
\renewcommand{\b}{\bcket{(}{)}}
\newcommand{\sqb}{\bcket{[}{]}}
\newcommand{\abs}{\bcket{\lvert}{\rvert}}

\renewcommand{\P}[1][]{\operatorname{P}_{#1}\b}
\newcommand{\Q}[1][]{\operatorname{Q}_{#1}\b}
\newcommand{\N}{\mathcal{N}\b}

\renewcommand{\H}{\mathbf{H}}

\newcommand{\y}{\mathbf{y}}
\newcommand{\F}{\mathbf{F}}

\newcommand{\x}{\mathbf{x}}

\newcommand{\w}{\mathbf{w}}

\renewcommand{\S}{\mathbf{\Sigma}}

\newcommand{\dd}[2][]{\frac{\partial #1}{\partial #2}}

\newcommand{\vmu}{\boldsymbol{\mu}}

\DeclareMathOperator*{\E}{\mathbb{E}}

\DeclareMathOperator*{\tr}{Tr}
\DeclareMathOperator*{\argmax}{arg\,max}

\newcommand{\Dkl}{\operatorname{D}_\text{KL}\b}
\newcommand{\tsum}{{\textstyle\sum}}
\newcommand{\g}{\mathbf{g}}

\title{Variational Laplace for Bayesian neural networks}

\author{%
  Ali Unlu,\\
  Department of Infomatics,\\
  University of Sussex,\\
  Brighton, UK\\
  \texttt{a.unlu@sussex.ac.uk}\\
  \And
  Laurence Aitchison,\\
  Department of Computer Science,\\
  University of Bristol,\\
  Bristol, UK\\
  \texttt{laurence.aitchison@gmail.com}
}

\begin{document}

\maketitle

\begin{abstract}
We develop variational Laplace for Bayesian neural networks (BNNs) which exploits a local approximation of the curvature of the likelihood to estimate the ELBO without the need for stochastic sampling of the neural-network weights.
The Variational Laplace objective is simple to evaluate, as it is (in essence) the log-likelihood, plus weight-decay, plus a squared-gradient regularizer.
Variational Laplace gave better test performance and expected calibration errors than maximum a-posteriori inference and standard sampling-based variational inference, despite using the same variational approximate posterior.
Finally, we emphasise care needed in benchmarking standard VI as there is a risk of stopping before the variance parameters have converged. 
We show that early-stopping can be avoided by increasing the learning rate for the variance parameters.\footnote{Anonymized code: anonymous.4open.science/r/fitr-FED4; MIT Licensed}
\end{abstract}


\section{Introduction}

Neural networks are increasingly being used in safety-critical settings such as self-driving cars \citep{bojarski2016end} and medical diagnosis \citep{amato2013artificial}.
In these settings, it is critical to be able to reason about uncertainty in the parameters of the network, for instance so that the system is able to call for additional human input when necessary \citep{mcallister2017concrete}.
Several approaches to Bayesian inference in neural networks are available, including stochastic gradient Langevin dynamics \citep{welling2011bayesian} Laplace's method \citep{azevedo1994laplace,mackay2003information,ritter2018scalable} and variational inference \citep{blundell2015weight,ober2020global}.

Here, we focus on combining the advantages of Laplace's method \citep{azevedo1994laplace,mackay2003information,ritter2018scalable} and variational inference \citep[VI; ][]{wainwright2008graphical}.
In particular, Laplace's method is very fast as it begins by finding a mode using a standard gradient descent procedure, then computes a local Gaussian approximate of the mode by performing a second-order Taylor expansion.
However, as the mode is discovered by standard gradient descent, it may be a narrow mode that generalises poorly \citep{neyshabur2017exploring}.
In contrast, variational inference \citep[VI;][]{blundell2015weight} is slower as it requires stochastic sampling of the weights, but that stochastic sampling forces it to find a broad, flat mode that presumably generalises better.
Here, we develop a new Variational Laplace (VL) method that combines the best of both worlds, giving a method that finds broad, flat modes even in the absence of the stochastic sampling.
The resulting objective is composed of the log-likelihood, standard weight-decay regularization and a squared-gradient regularizer, which is weighted by the variance of the approximate posterior.
VL displayed improved performance over VI and MAP on standard benchmark tasks.

\section{Background}
\subsection{Variational inference (VI) for Bayesian neural networks}

To perform Variational Inference for neural networks, we follow the usual approach \citep{hinton1993keeping,blundell2015weight}, in using independent Gaussian priors, $\operatorname{P}$ and approximate posteriors $\operatorname{Q}$ for all parameters, $\w$,
\begin{align}
  \label{eq:def:prior_ap}
  \P{w_\lambda} &= \N{w_\lambda; 0, s_\lambda^2} \\
  \Q{w_\lambda} &= \N{w_\lambda; \mu_\lambda, \sigma_\lambda^2} && \text{ equivalently } &
  \Q{\w} &= \N{\w; \vmu, \S},
\end{align}
where $\mu_\lambda$ and $\sigma_\lambda^2$ are learned parameters of the approximate posterior, and where $\S$ is a diagonal matrix, with $\Sigma_{\lambda \lambda} = \sigma_\lambda^2$. 
We fit the approximate posterior by optimizing the evidence lower bound objective (ELBO) with respect to parameters of the variational posterior, $\mu_\lambda$ and $\sigma_\lambda^2$, 
\begin{equation}
  \label{eq:def:elbo}
  \mathcal{L}_\text{VI} = \E_{\Q{\w}}\biggl[\log \P{\y| \x, \w}+ \beta \tsum_\lambda \log \frac{\log \P{w_\lambda}}{\log \Q{w_\lambda}}\biggr].
\end{equation}
Here, $\x$ is all training inputs, $\y$ is all training outputs, and $\beta$ is the tempering parameter which is $1$ for a close approximation to Bayesian inference, but is often set to smaller values to ``temper'' the posterior, which often improves empirical performance \citep{huang2018improving,wenzel2020good} and has theoretical justification as accounting for the data-curation process \citep{aitchison2020statistical}.

We need to optimize the expectation in Eq.~\eqref{eq:def:elbo} with respect to the parameters of $\Q{\w}$, the distribution over which the expectation is taken.
To perform this optimization efficiently, we therefore use the reparameterisation trick \citep{kingma2013auto,rezende2014stochastic,blundell2015weight} --- we write $\w$ in terms of $\boldsymbol{\epsilon}$,
\begin{align}
  w_\lambda(\epsilon_\lambda) &= \mu_\lambda + \sigma_\lambda \epsilon_\lambda
\end{align}
where $\epsilon_\lambda \sim \N{0, 1}$.
Thus, the ELBO can be written as an expectation over $\boldsymbol{\epsilon}$,
\begin{equation}
  \label{eq:def:reparam}
  \mathcal{L}_\text{VI} = \E_{\boldsymbol{\epsilon}}\biggl[\log \P{\y| \x, \w(\boldsymbol{\epsilon})}+ \beta \tsum_\lambda \log \frac{\log \P{w_\lambda(\epsilon_\lambda)}}{\log \Q{w_\lambda(\epsilon_\lambda)}}\biggr].
\end{equation}
where the distribution over $\boldsymbol{\epsilon}$ is now fixed.
Critically, now the expected gradient of the term inside the expectation is equal to the gradient of $\mathcal{L}_\text{VI}$, so we can use samples of $\boldsymbol{\epsilon}$ to estimate the expectation.

\subsection{Laplace's method}

Laplace's method \citep{azevedo1994laplace,mackay2003information,ritter2018scalable} first finds a mode by doing gradient ascent on the log-joint,
\begin{align}
  \w^* &= \argmax_{\w} \sqb{\log \P{\y| \x, \w} + \log \P{\w}}
\end{align}
and uses a Gaussian approximate posterior around that mode,
\begin{align}
  \Q{\w} &= \N{\w; \w^*, -\H^{-1}(\w^*)}
\end{align}
where $\H(\w^*)$ is Hessian of the log-joint at $\w^*$.

%

\section{Related work}

There is past work on Variational Laplace \citep{friston2007variational,daunizeau2009variational,daunizeau2017variational}, which learns the mean parameters, $\boldsymbol{\mu}$, of a Gaussian approximate posterior,
\begin{align}
  \Q[\vmu]{\w} &= \N{\w; \boldsymbol{\mu}, -\H^{-1}(\vmu)}
\end{align}
and obtains the covariance matrix as a function of the mean parameters using the Hessian, as in Laplace's method.
However, instead of taking the approximation to be centered around a MAP solution, $\w^*$, they take the approximate posterior to be centered on learned mean parameters, $\vmu$. 
Importantly, they simplify the ELBO by substituting this approximate posterior into Eq.~\ref{eq:def:elbo}, and approximating the log-joint using its Taylor series expansion. 
Ultimately they obtain,
\begin{equation}
  \label{eq:vl:past}
  \mathcal{L}_\text{VI} \approx \log \P{\y| \w{=}\vmu, \x} + \log \P{\w{=}\vmu} 
    - \tfrac{1}{2} \log \abs{\H(\vmu)} + \text{const}.
\end{equation}
However, there are two problems with this approach when applied to neural networks.
First, the algebraic manipulations required to derive Eq.~\eqref{eq:vl:past} require the full $N \times N$ Hessian, $\H(\vmu)$, for all $N$ parameters, and neural networks have too many parameters for this to be feasible.
Second, the $\log \abs{\H(\vmu)}$ term in Eq.~\eqref{eq:vl:past} cannot be minibatched, as we need the full sum over minibatches inside the $\log$ to compute the Hessian,
\begin{align}
  \log \abs{\H(\vmu)} &= \log \abs{\tsum_j \H_j(\vmu)},
\end{align}
where $\H_j(\vmu)$ is the contribution to the Hessian from an individual minibatch.
Due to these issues, past Variational Laplace methods did not scale to large neural networks.


An alternative deterministic approach to variational inference in Bayesian neural networks, approximates the distribution over activities induced by stochasticity in the weights \citep{wu2018deterministic}.
Unfortunately, it is important to capture the covariance over features induced by stochasticity in the weights.
In fully connected networks, this is feasible, as we usually have a small number of features at each layer.
However, in convolutional networks, we have a large number of features, $\texttt{channels} \times \texttt{height} \times \texttt{width}$. 
In the lower layers of a ResNet, we may have $64$ channels and a $32 \times 32$ feature map, resulting in $64\times32^2=65,536$ features and a $65,536 \times 65,536$ covariance matrix. 
These scalability issues prevented them from applying their approach to convolutional networks.
In contrast, our approach is highly scalable and readily applicable to the convolutional setting.

\citet{ritter2018scalable} and \citet{mackay1992practical} used Laplace's method in Bayesian neural networks, by first finding the mode by doing gradient ascent on the log-joint probability, and expanding around that mode.
As usual for Laplace's method, they risk finding a narrow mode that generalises poorly.
In contrast, we find a mode using an approximation to the ELBO that takes the curvature into account and hence is biased towards broad, flat modes that presumably generalise better.

Finally, our approach will eventually give a squared-gradient regularizer that is similar to those discovered in past work \citep{barrett2020implicit,smith2021origin}. 
This work found a slightly different squared-gradient regularizer has a connection to gradient descent, in that approximation errors due to finite-step sizes in gradient-descent imply an effective squared gradient regularization.
The similarity of our objectives raises profound questions about the extent to which gradient descent can be said to perform Bayesian inference.
That said there are two key differences.
First, our method uses the Fisher, (i.e.\ the gradients for data sampled from the model) whereas their approach uses the empirical Fisher, (i.e.\ gradients for the observed data) to form the squared gradient regularizer \citep{kunstner2019limitations}.
Second, our approach gives a principled method to learn a separate weighting for the squared-gradient for each parameter, whereas the connection to SGD forces \citet{barrett2020implicit} to use a uniform weighting across all parameters.

\section{Methods} 

To combine the best of VI and Laplace's method, we begin by noting that the ELBO can be rewritten in terms of the KL divergence between the prior and approximate posterior,
\begin{equation}
  \label{eq:def:elbo:kl}
  \mathcal{L}_\text{VI} = \E_{\Q{\w}}\sqb{\log \P{\y| \x, \w}}
  -\beta \tsum_\lambda \Dkl{\Q{w_\lambda}|| \P{w_\lambda}},
\end{equation}
where the KL-divergence can be evaluated analytically,
\begin{align}
  \label{eq:def:kl}
  D_\text{KL}\b{\Q{w_\lambda}|| \P{w_\lambda}} 
  &= \frac{1}{2} \b{\frac{\sigma_\lambda^2 + \mu_\lambda^2}{s_\lambda^2} - 1 + \log \frac{s_\lambda^2}{\sigma_\lambda^2} }.
\end{align}
As such, the only term we need to approximate is the expected log-likelihood.

To approximate the expectation, we begin by taking a second-order Taylor series expansion of the log-likelihood around the current setting of the mean parameters, $\vmu$,
\begin{multline}
  \E_{\Q{\w}}\big[\log \P{\y| \x, \w}\big] \approx \log \P{\y| \x, \w{=}\vmu}
  + \E_{\Q{\w}}\sqb{\tsum_{j=1}^B \g_j^T \b{\w - \vmu}}\\
  + \E_{\Q{\w}}\sqb{\tfrac{1}{2} \b{\w - \vmu}^T \H \b{\w - \vmu}}
\end{multline}
where $B$ is the number of minibatches, $\g_j$ is the gradient for minibatch $j$ and $\H$ is the Hessian for the full dataset,
\begin{align}
  \label{eq:def:g}
  g_{j; \lambda} &= \dd{w_\lambda} \sqb{\log \P{\y_j| \x_j, \w}} \\ 
  H_{\lambda, \nu} &= \frac{\log \P{\y| \x, \w}}{\partial w_\lambda \partial w_\nu}.
\end{align}
Here, $\x$ and $\y$ are the the inputs and outputs for the full dataset, whereas $\x_j$ and $\y_j$ are the inputs and outputs for minibatch $j$.
Now we consider the expectation of each of these terms under the approximate posterior, $\Q{\w}$.  The first term is constant and independent of $\w$.
The second (linear) term is zero, because the expectation of $\b{\w - \vmu}$ under the approximate posterior is zero
\begin{align}
  \E_{\Q{\w}}\sqb{\g_j^T \b{\w - \vmu}} &= \g_j^T \E_{\Q{\w}}\sqb{\b{\w - \vmu}} = 0.
\end{align}
The third (quadratic) term might at first appear difficult to evaluate because it involves $\H$, the $N \times N$ matrix of second derivatives, where $N$ is the number of parameters in the model.
However, using properties of the trace, and noting that the expectation of $\b{\w - \vmu} \b{\w - \vmu}^T$  is the covariance of the approximate posterior we obtain,
\begin{align}
  \E_{\Q{\w}}\Big[\tfrac{1}{2} (\w &- \vmu)^T  \H (\w - \vmu)\Big] =\E_{\Q{\w}}\sqb{\tfrac{1}{2} \tr\b{\H \b{\w - \vmu} \b{\w - \vmu}^T}} = \tfrac{1}{2} \tr\b{\H \S}
\end{align}
writing the trace in index notation, and substituting for the (diagonal) posterior covariance, $\S$,
\begin{equation}
  \tfrac{1}{2} \tr\b{\H \S}= \tfrac{1}{2} \sum_{\lambda \nu} H_{\lambda \nu} \Sigma_{\lambda \nu}
  = \tfrac{1}{2} \sum_{\lambda} H_{\lambda \lambda} \sigma_\lambda^2.
\end{equation}
Thus, our first approximation of the expected log-likelihood is,
\begin{equation}
  \label{eq:exp_like_hess}
  \E_{\Q{\w}}\sqb{\log \P{\y| \x, \w}} \approx
  \log \P{\y| \x, \w{=}\vmu} + \tfrac{1}{2} \tsum_\lambda \sigma_\lambda^2 H_{\lambda\lambda},
\end{equation}
and substituting this into Eq.~\eqref{eq:def:elbo:kl} gives,
\begin{multline}
  \label{eq:exp_elbo_hess}
  \mathcal{L}_\text{VI} \approx \mathcal{L}_\text{VL(H)} = \log \P{\y| \x, \w{=}\vmu} + \tfrac{1}{2} \tsum_{\lambda} \sigma_\lambda^2 H_{\lambda \lambda}
    - \beta \tsum_\lambda \Dkl{\Q{w_\lambda}|| \P{w_\lambda}}.
\end{multline}
This resolves most of the issues with the original Variational Laplace method: it requires only the diagonal of the Hessian, it can be minibatched and it does not blow up if $H_{\lambda \lambda}$ is zero.

\subsection{Pathological optima when using the Hessian}

However, a new issue arises: $H_{\lambda \lambda}$ is usually negative, in which case the approximation in Eq.~\eqref{eq:exp_elbo_hess} can be expected to work well.
However there is nothing to stop $H_{\lambda \lambda}$ from becoming positive. 
Usually if we e.g.\ took the log-determinant of the negative Hessian, this would immediately break the optimization process (as we would be taking the logarithm of a negative number).
However, in our context, there is no immediate issue as Eq.~\eqref{eq:exp_elbo_hess} takes on a well-defined value even when one or more $H_{\lambda \lambda}$'s are positive.
That said, we rapidly encounter similar issues as we get pathological optimal values of $\sigma_\lambda^2$.
In particular, picking out the terms in the objective that depend on $\sigma_\lambda^2$, absorbing the other terms into the constant, and taking $\beta=1$ for simplicity, we have
\begin{align}
  \mathcal{L}_\text{VL(H)} = \tfrac{1}{2} \tsum_{\lambda} \b{- \b{\tfrac{1}{s_\lambda^2} - H_{\lambda \lambda}} \sigma_\lambda^2 + \log \sigma_\lambda^2} + \text{const}.
\end{align}
Thus, the gradient wrt a single variance parameter is,
\begin{align}
  \dd{\sigma_\lambda^2} \mathcal{L}_\text{VL(H)} = \tfrac{1}{2} \b{- \b{\tfrac{1}{s_\lambda^2} - H_{\lambda \lambda}} + \tfrac{1}{\sigma_\lambda^2}}.
\end{align}
In the typical case, $H_{\lambda \lambda}$ is negative so $\b{\tfrac{1}{s_\lambda^2} - H_{\lambda \lambda}}$ is positive, and we can find the optimum by solving for the value of $\sigma_\lambda^2$ where the gradient is zero,
\begin{align}
  \label{eq:sigma_opt}
  \sigma_\lambda^2 &= \frac{1}{\tfrac{1}{s_\lambda^2} -H _{\lambda \lambda}}.
\end{align}
However, if $H_{\lambda \lambda}$ is positive and sufficiently large, $H_{\lambda \lambda} > \tfrac{1}{s_\lambda^2}$, then $\b{\tfrac{1}{s_\lambda^2} - H_{\lambda \lambda}}$ becomes negative, and not only is the mode in Eq.~\eqref{eq:sigma_opt} undefined, but the gradient is always positive,
\begin{align}
  0 < \dd{\sigma_\lambda^2} \mathcal{L}_\text{VL(H)} = \tfrac{1}{2} \b{- \b{\tfrac{1}{s_\lambda^2} - H_{\lambda \lambda}} + \tfrac{1}{\sigma_\lambda^2}}.
\end{align}
as both terms in the sum: $- \b{\tfrac{1}{s_\lambda^2} - H_{\lambda \lambda}}$ and $\tfrac{1}{\sigma_\lambda^2}$ are positive.
As such, when $H_{\lambda \lambda} > \tfrac{1}{s_\lambda^2}$, the variance, $\sigma^2_\lambda$ grows without bound.

\subsection{Avoiding pathologies with the Fisher}

To avoid pathologies arising from the fact that the Hessian is not necessarily negative definite, a common approach is to approximate the Hessian using the Fisher Information matrix,
\begin{align}
  \label{eq:approx_bound}
  - \H \approx \F &= \sum_{j=1}^B \E_{\P{\mathbf{\tilde{y}}_j| \x_j, \w=\vmu}}\sqb{\mathbf{\tilde{g}}_j(\mathbf{\tilde{y}}_j) \mathbf{\tilde{g}}_j^T(\mathbf{\tilde{y}}_j)}.
\end{align}
Importantly, $\mathbf{\tilde{g}}$ is the gradient of the log-likelihood for data sampled from the model, $\mathbf{\tilde{y}}_j$, \textit{not} for the true data,
\begin{align}
  \label{eq:def:gt}
  \tilde{g}_{j; \lambda}(\mathbf{\tilde{y}}_j) &= \dd{w_\lambda} \sqb{\log \P{\mathbf{\tilde{y}}_j| \x_j, \w}}.
\end{align}
This gives us the Fisher, which is a commonly used and well-understood approximation to the Hessian \citep{kunstner2019limitations}.
Importantly, this contrasts with the empirical Fisher \citep{kunstner2019limitations}, which uses the gradient conditioned on the actual data (and not data sampled from the model),
\begin{align}
  \F_\text{emp} &= \sum_{j=1}^B \g_j \g_j^T,
\end{align}
which is problematic, because there is a large rank-1 component in the direction of the mean gradient, which disrupts the estimated matrix specifically in the direction of interest for problems such as optimization \citep{kunstner2019limitations}.


\begin{figure*}[t]
\centering
  \includegraphics[width=5.5in]{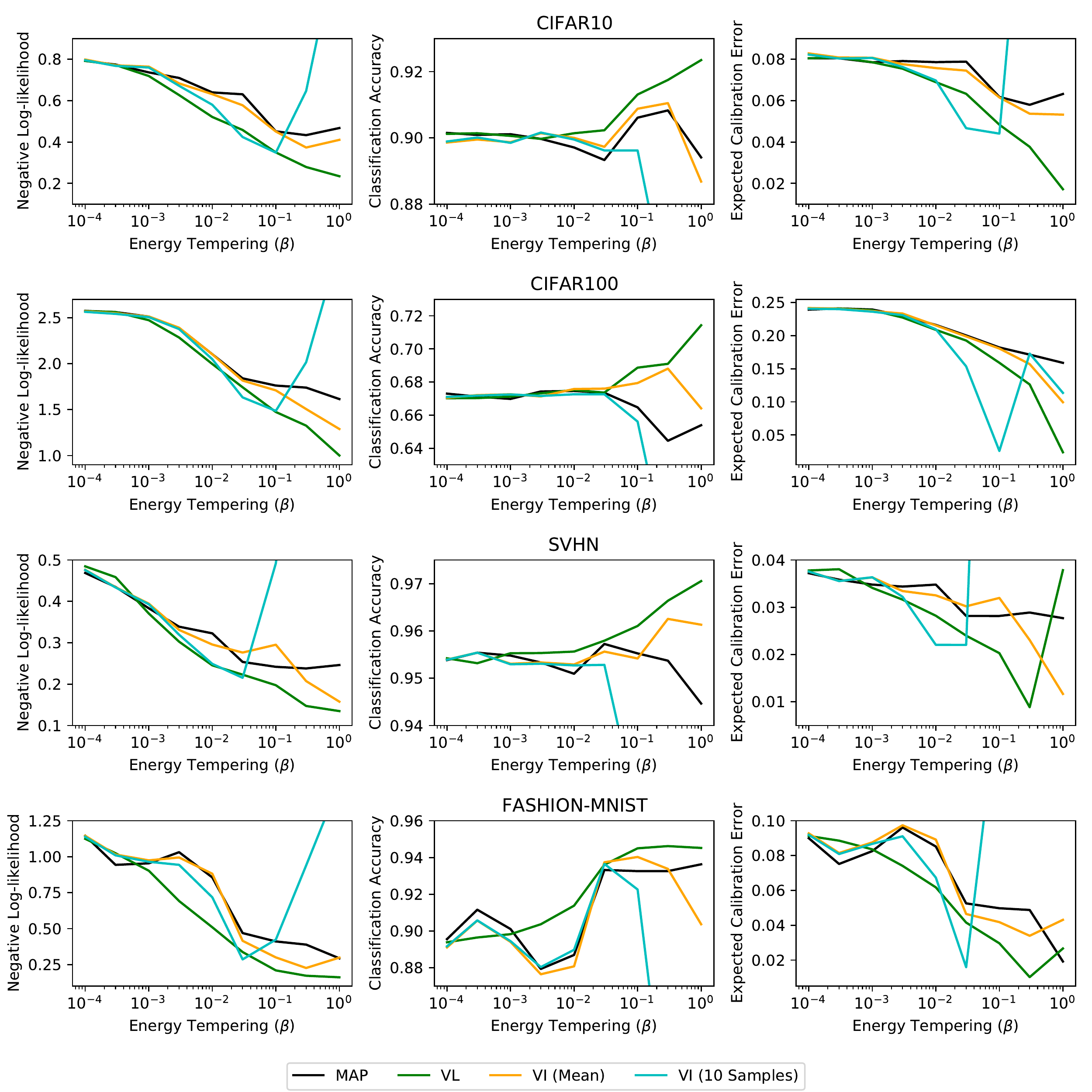}
  \caption{
    Training a PreactResNet-18 on various datasets, displaying the test accuracy, test log-likelihood and expected calibration error (ECE) \citep{naeini2015obtaining,guo2017calibration} for CIFAR-10, CIFAR-100, SVHN and fashion MNIST.
    \label{fig:fixed}
  }
\end{figure*}


\begin{table}[t]
\begin{center}
\setlength\tabcolsep{2pt}
\caption{Best values test NLL, test accuracy and ECE for a variety of datasets as we use different values of the tempering parameter, $\beta$ \vspace{2pt}}
\begin{tabular}{ccccc}
\toprule
dataset & method & test NLL & test acc. & ECE\\
\midrule
         & VL  & \textbf{0.23} & \textbf{92.4\%} & \textbf{0.017} \\
\multirow{ 2}{*}{CIFAR-10} & VI (Mean) & 0.37 & 91.1\% & 0.053\\
         & VI (10 Samples) & 0.35 & 90.2\% & 0.044\\
         & MAP & 0.43 & 90.8\% & 0.058\\
\midrule
          &VL  & \textbf{1.00} & \textbf{71.4\%} & \textbf{0.024} \\
\multirow{ 2}{*}{CIFAR-100} &VI (Mean) & 1.29 & 68.8\% & 0.100\\
          & VI (10 Samples) & 1.49 & 67.3\% & 0.026\\
          &MAP & 1.61 & 67.5\% & 0.159\\
\midrule
     &VL  & \textbf{0.14} & \textbf{97.1\%} & \textbf{0.009} \\
\multirow{ 2}{*}{SVHN} &VI (Mean) & 0.16 & 96.3\% & 0.012\\
     &VI (10 Samples) & 0.22 & 95.5\% & 0.022\\
     &MAP & 0.24 & 95.7\% & 0.028\\
\midrule
              &VL  & \textbf{0.16} & \textbf{94.6\%} & \textbf{0.010} \\
\multirow{ 2}{*}{Fashion MNIST} &VI (Mean) & 0.23 & 94.0\% & 0.034\\
              & VI (10 Samples) & 0.29 & 93.6\% & 0.016\\
              &MAP & 0.29 & 93.6\% & 0.096\\
\bottomrule
\end{tabular}
\label{tab}
\end{center}
\end{table}

\begin{figure*}
  \centering
  \includegraphics[width=0.95\textwidth]{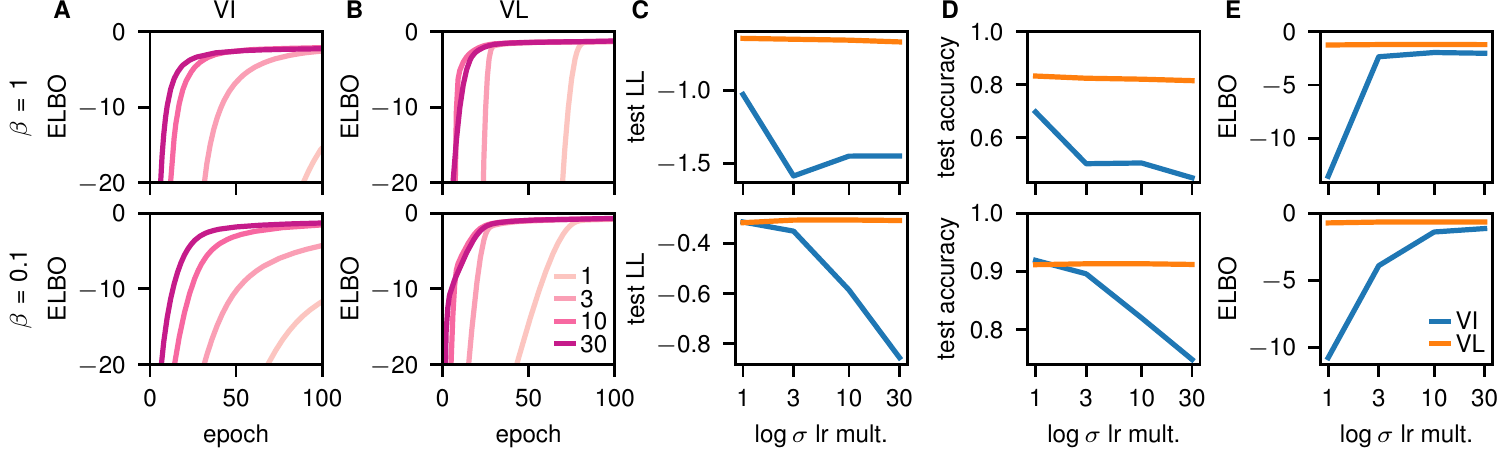}
  \caption{
    Analysis of early stopping in VI and VL.
    The first row is untempered ($\beta=1$), and the second row is tempered ($\beta=0.1$).
    \textbf{A} ELBO over epochs 0-100 (with the highest initial learning rate) for VI. 
    Different lines correspond to networks with learning rate multipliers for $\log \sigma_\lambda$ of 1, 3, 10 and 30.
    \textbf{B} As \textbf{A}, but for VL.
    \textbf{CDE} Final test-log-likelihood (\textbf{C}), test accuracy (\textbf{D}) and ELBO (\textbf{E}) after 200 epochs for different learning rate multipliers.
    \label{fig:early_stopping}
  }
\end{figure*}
Using the Fisher Information (Eq.~\ref{eq:approx_bound}) in Eq.~\eqref{eq:exp_like_hess}, we obtain an approximate expected log-likelihood,
\begin{equation}
  \E_{\Q{\w}}\sqb{\log \P{\y| \x, \w}} \approx \log \P{\y| \x, \w{=}\vmu} 
  - \tfrac{1}{2} \tsum_\lambda \sigma_\lambda^2 \tsum_{j=1}^B \tilde{g}_{j; \lambda}^2.
\end{equation}
Substituting this into Eq.~\eqref{eq:def:elbo:kl} gives us the final VL objective, $\mathcal{L}_\text{VL}$, which is an approximation to the ELBO,
\begin{equation}
  \mathcal{L}_\text{VI} \approx \mathcal{L}_\text{VL} = \log \P{\y| \x, \w{=}\vmu} - \tfrac{1}{2} \tsum_{\lambda} \sigma_\lambda^2 \tsum_{j=1}^B \tilde{g}_{j; \lambda}^2 
  - \beta \tsum_\lambda \Dkl{\Q{w_\lambda}|| \P{w_\lambda}}.
\end{equation}
In practice, we typically take the objective for a minibatch, divided by the number of datapoints in a minibatch, $S$,
\begin{align}
  \tfrac{1}{S} \mathcal{L}_{\text{VL}; j} = \tfrac{1}{S} \log \P{\y_j| \x_j, \w{=}\vmu}
  - \tfrac{S}{2} \tsum_{\lambda} \sigma_\lambda^2 \b{\tfrac{1}{S} \tilde{g}_{j; \lambda}}^2 
  - \tfrac{\beta}{2SB} \tsum_\lambda \b{\tfrac{\sigma_\lambda^2 + \mu_\lambda^2}{s_\lambda^2} - 1 + \log \tfrac{s_\lambda^2}{\sigma_\lambda^2} },
\end{align}
where $\b{\tfrac{1}{S} \tilde{g}_{j; \lambda}}$ are the gradients of the log-likelihood for the minibatch averaged across datapoints, i.e.\ the gradient of $\tfrac{1}{S} \log \P{\mathbf{\tilde{y}}_j| \x_j, \w{=}\vmu}$.
Remember $B$ is the number of minibatches so $SB$ is the total number of training datapoints.

\subsection{Constraints on the network architecture}

Importantly, here the regularizer is the squared gradient of the loss with respect to the parameters.
As such, computing the loss implicitly involves a second-derivative of the log-likelihood, and we therefore cannot use piecewise linear activation functions such as ReLU, which have pathological second derivatives.
In particular, the second derivative has a delta-function ``spike'' at zero,
\begin{align}
  \frac{d^2}{dx} \phi(x) &= \frac{d}{dx} \sqb{\frac{d}{dx}\phi(x)} = \frac{d}{dx} \Theta(x) = \delta(x)
\end{align}
where $\phi$ is the relu nonlinearity, $\Theta(x)$ is the Heaviside step function which is zero for $x<0$ and one for $0<x$, and $\delta(x)$ is the Dirac delta function.
As the function is almost never evaluated at exactly zero, it is not possible to sensibly take into account the contribution of the infinitely high spike in the second derivative at zero.
Interestingly, this issue is very similar to the one that turns up when differentiating step (i.e.\ $\Theta(x)$) activations --- the derivative is well-defined and zero almost everywhere --- the issue is there are delta-function spikes in the gradient at zero that gradient descent cannot reasonably work with.
Instead, we used a softplus activation function, but any activation with well-behaved second derivatives is admissible.

\section{Results}

We compared MAP, VI and our method (VL) on four different datasets (CIFAR-10, CIFAR-100 \citealp{krizhevsky2009learning}, SVHN \citealp{netzer2011reading} and fashion-MNIST \citealp{xiao2017fashion} MIT Licensed) using a PreactResNets-18 \citep{he2016identity} with an initial learning rate of 1E-4, which decreased by a factor of 10 after 100 and 150 epochs and a batch size of 128 with all the other optimizer hyperparameters set to their default values. 
We tried two variants of variational inference: evaluating test-performance using the mean network, VI (mean), and evaluating test performance by drawing 10 samples from the approximate posterior, VI (sampled).
We swept across different degrees of posterior tempering, $\beta$.
Using $\beta<1$ is normatively justified in the Bayesian framework as accounting for the effect of data curation \citep{aitchison2020statistical}.
For many values of $\beta$ VL gave better test accuracies, test log-likelihoods and expected calibration errors \citep{naeini2015obtaining,guo2017calibration} than VI or MAP inference (Fig.~\ref{fig:fixed}).
Importantly though, for the optimal value of $\beta$, VL almost always gave better performance on these metrics (Table~\ref{tab}).
These experiments took $\sim 480$ GPU hours, and were run on a mixture of nVidia 1080 and 2080 GPUs in an internal cluster.

\begin{table}[t]
\begin{center}
\setlength\tabcolsep{2pt}
\caption{Time per epoch for different methods on CIFAR-10 \vspace{2pt}}
\begin{tabular}{cc}
\toprule
method & time per epoch (s)\\
\midrule
VL  & 114.9\\
VI  & 43.2\\
MAP & 41.8\\
\bottomrule
\end{tabular}
\label{tab:time}
\end{center}
\end{table}
The runtime of the methods is listed in Table~\ref{tab:time}.
VL is around a factor of 3 slower than either VI or VL due to the need to compute second-derivatives, it is still eminently feasible, especially in comparison to past methods to deterministic variational inference that have fundamental difficulties in scaling to convolutional networks \citep{wu2018deterministic}.
Furthermore, we did not find that increasing the number of epochs improved performance either for VI or MAP as we are already training to convergence.

\subsection{Early-stopping and poor performance in VI}
\label{app:early_stop}

Before performing comparisons where we learn the approximate posterior variance, it is important to understand the pitfalls when optimizing variational Bayesian neural networks using adaptive optimizers such as Adam.
In particular, there is a strong danger of stopping the optimization before the variances have converged.
To illustrate this risk, note that Adam \citep{kingma2014adam} updates take the form,
\begin{align}
  \Delta \theta &= \eta \frac{m}{\sqrt{v} + \epsilon}
\end{align}
where $\eta$ is the learning rate, $m$ is an unbiased estimator of the mean gradient, $\langle g \rangle$, $v$ is an unbiased estimator of the squared gradient, $\langle g^2 \rangle$, and $\epsilon$ is a small positive constant to avoid divide-by-zero.
The magnitude of the updates, $\abs{\Delta \theta}$, is maximized by having exactly the same gradient on each step, in which case, neglecting $\epsilon$, we have $\abs{\Delta \theta} = \eta$.
As such, with a learning rate of $\eta=10^{-4}$, a training set of $50,000$ and a batch size of $128$ parameters can move at most $50,000 / 128 \times 10^{-4} \approx 0.04$ per epoch.
Doing 100 epochs at this learning rate, a parameter can change by at most $4$ over the $100$ epochs before the first learning rate step.

This is fine for the weights, which typically have very small values.
However, the underlying parameters used for the variances typically take on larger values.
In our case, we will use $\log \sigma_\lambda$ as the parameter, and initialize it to 3 less than the prior standard deviation, $\log s_\lambda - 3$.
To ensure reasonable convergence, $\log \sigma_\lambda$ should be able to revert back to the prior, implying that it must be able to change by at least 3 during the course of training. 
Unfortunately, 3 is very close to the maximum possible change of 4, raising the possibility that the variance parameters will not actually converge.
To check whether early-stopping was indeed an issue, we plotted the (tempered) ELBO for VI (Fig.~\ref{fig:early_stopping}A) and VL (Fig.~\ref{fig:early_stopping}B).
For VI (Fig.~\ref{fig:early_stopping}A) with the standard setup (lightest line with a learning rate multiplier of $1$), the ELBO clearly has not converged at 100 epochs, indicating early-stopping.
Notably, this was still an issue with VL (Fig.~\ref{fig:early_stopping}B), especially if we were to train for fewer epochs.
However, the effect is smaller for VL, which may be because the gradients are more consistent as it does not sample the weights.
These issues can be rectified by increasing the learning rate specifically for the $\log \sigma_\lambda$ parameters (darker lines).


We then plotted the test log-likelihood (Fig.~\ref{fig:early_stopping}C), test accuracy (Fig.~\ref{fig:early_stopping}D) and ELBO (Fig.~\ref{fig:early_stopping}E) against the learning rate multiplier.
Again, the performance for VL (orange) was reasonably robust to changes in the learning rate multiplier.
However, the performance of VI (blue) was very sensitive to the multiplier: as the multiplier increased, test performance fell but the ELBO rose.
As we ultimately care about test performance, these results would suggest that we should use the lowest multiplier (1), and accept the possibility of early-stopping.
That may be a perfectly good choice in many cases.
However, VI is supposed to be an approximate Bayesian method, and using an alternative form for the ELBO,
\begin{align}
  \mathcal{L}_\text{VI} &= \log \P{\y| \x} - D_\text{KL}\b{\Q{\w} || \P{\w| \y, \x}},
\end{align}
we can see that the ELBO measures KL-divergence between the true and approximate posterior, and hence the quality of our approximate Bayesian inference.
As such, very poor ELBOs imply that the KL-divergence between the true and approximate posterior is very large, and hence the ``approximate posterior'' is no longer actually approximating the true posterior.
As such, if we are to retain a Bayesian interpretation of VI, we need to use larger learning rate multipliers which give better values for the ELBO (Fig.~\ref{fig:early_stopping}E).
However, in doing that, we get worse test performance (Fig.~\ref{fig:early_stopping}CD).
This conflict between approximate posterior quality and test performance is very problematic: the Bayesian framework would suggest that as Bayesian inference becomes more accurate, performance should improve, whereas for VI, performance gets worse.
Concretely, by initializing $\log \sigma_\lambda$ to a small value and then early-stopping, we leave $\log \sigma_\lambda$ at a small value through training, in which case VI becomes equivalent to MAP inference with a negligibly small amount of noise added to the weights. 
We would therefore expect early-stopped VI to behave (and be) very similar to MAP inference.

In subsequent experiments, we chose to use a learning rate multiplier of 10, as this largely eliminated early-stopping (though see VI with $\beta=0.1$; Fig.~\ref{fig:early_stopping}E).

\section{Conclusions}

We gave a novel Variational Laplace approach to inference in Bayesian neural networks which combines the best of previous approaches based on Variational Inference and Laplace's Method.
This method gave excellent empirical performance compared to VI.

No negative social impacts are anticipated as this is largely theoretical work.


\bibliographystyle{icml2021}
\bibliography{refs}

\end{document}